\documentclass[twocolumn, switch]{article}

\usepackage{preprint}

\usepackage{amsmath, amsthm, amssymb, amsfonts}

\usepackage[numbers,square]{natbib}
\usepackage[utf8]{inputenc}
\usepackage[T1]{fontenc}
\usepackage{xcolor}
\usepackage[colorlinks = true,
            linkcolor = purple,
            urlcolor  = black,
            citecolor = cyan,
            anchorcolor = black]{hyperref}
\usepackage{booktabs}
\usepackage{nicefrac}
\usepackage{microtype}
\usepackage{lineno}
\usepackage{float}
\usepackage{graphicx}
\usepackage{siunitx}
\usepackage{caption}
\usepackage{placeins}
\usepackage{bbm}
\usepackage{colortbl}
\usepackage{makecell}

\usepackage{titlesec}
\titlespacing\section{0pt}{12pt plus 3pt minus 3pt}{1pt plus 1pt minus 1pt}
\titlespacing\subsection{0pt}{10pt plus 3pt minus 3pt}{1pt plus 1pt minus 1pt}
\titlespacing\subsubsection{0pt}{8pt plus 3pt minus 3pt}{1pt plus 1pt minus 1pt}

\definecolor{lightblue}{RGB}{219, 233, 246}
\definecolor{mediumblue}{RGB}{91, 155, 213}
\definecolor{lightred}{RGB}{255, 230, 230}
\definecolor{lightgreen}{RGB}{230, 255, 230}
\definecolor{headerblue}{RGB}{68, 114, 196}

\newcommand{\comparecell}[3]{%
\ifnum\numexpr\pdfstrcmp{#1}{#2}>0
\cellcolor{lightgreen!80}#3
\else
\ifnum\numexpr\pdfstrcmp{#1}{#2}<0
\cellcolor{lightred!80}#3
\else
#3
\fi
\fi
}

\title{ShED-HD: A Shannon Entropy Distribution Framework for Lightweight Hallucination Detection on Edge Devices}

\usepackage{authblk}

\author[1]{Aneesh Vathul}
\author[1]{Daniel Lee}
\author[1]{Sheryl Chen}
\author[1]{Arthi Tasmia}

\affil[1]{Stanford University\\
\texttt{avathul@stanford.edu}, \texttt{leedan@stanford.edu}, \texttt{sherylch@stanford.edu}, \texttt{arthi26@stanford.edu}}

\begin{document}

\twocolumn[
  \begin{@twocolumnfalse}
  
\maketitle

\begin{abstract}
Large Language Models (LLMs) have demonstrated impressive capabilities on a broad array of NLP tasks, but their tendency to produce hallucinations—plausible-sounding but factually incorrect content—poses severe challenges in high-stakes domains. Existing hallucination detection methods either bear the computational cost of multiple inference passes or sacrifice accuracy for efficiency with single-pass approaches, neither of which is ideal in resource-constrained environments such as edge devices. We propose the Shannon Entropy Distribution Hallucination Detector (ShED-HD), a novel hallucination detection framework that bridges this gap by classifying sequence-level entropy patterns using a lightweight BiLSTM architecture with single-headed attention. In contrast to prior approaches, ShED-HD efficiently detects distinctive uncertainty patterns across entire output sequences, preserving contextual awareness. Through in-depth evaluation on three datasets (BioASQ, TriviaQA, and Jeopardy Questions), we show that ShED-HD significantly outperforms other computationally efficient approaches in the out-of-distribution setting, while achieving comparable performance in the in-distribution setting. ShED-HD facilitates hallucination detection that is low-cost, accurate, and generalizable, improving the credibility of content generated by LLMs in resource-constrained environments where trustworthy AI functionality is crucial.
\end{abstract}

\vspace{0.35cm}

  \end{@twocolumnfalse}
]

\section{Introduction}
\label{sec:intro}
Large Language Models (LLMs) have achieved unmatched performance on a wide range of natural language processing tasks, including text summarization, machine translation, question answering, and conversational agent generation \cite{zellers2020defendingneuralfakenews}. Such advances have made it possible for LLMs to be applied in many practical applications, particularly in high-stakes domains such as healthcare, finance, law practice, and scientific research. Applying LLMs in such fields requires that they produce reliable output, incorrect or misleading information can significantly impact end users, who are becoming increasingly reliant on these tools for critical decisions. Possibly the most significant challenge in deploying LLMs is their tendency to produce hallucinations—logically sounding yet factually false content. Hallucinations may consist of any type of nonfactual material generated by LLMs. Such hallucinations compromise the trustworthiness of LLM responses and limit their use in high-stakes scenarios such as education, medical information retrieval, and legal recommendation systems.

Recently, large language models that are optimized for edge devices via smaller architectures have also gained significant traction. This makes the task of reliable hallucination detection in resource-constrained environments increasingly important. Lightweight hallucination detection is essential for edge devices because of their limited processing power and memory, which makes running sophisticated, compute-intensive verification algorithms impractical. Moreover, many edge use cases (such as autonomous vehicles or medical devices) occur in environments where internet connectivity is unreliable, making it impossible to offload verification to cloud servers or internet-based retrieval methods. This further highlights the need for on-device solutions that can operate independently of external infrastructure. A robust, self-contained hallucination detection system would allow LLMs that operate on these devices to preserve their reliability, ensuring that they output accurate information even under bandwidth-limited or isolated conditions.

Several techniques for hallucination detection without the use of external sources have been explored. For example, multi-pass approaches to hallucination detection, such as self-sampling \cite{wang2023selfconsistencyimproveschainthought} and semantic entropy \cite{farquharSemanticEntropy}, operate by generating a set of varied responses to a question and analyzing the variation in the outputs. The assumption behind this approach is that when an LLM is knowledgeable about a subject with confidence, it will consistently generate similar responses. Conversely, when the model is uncertain, its outputs are inconsistent, indicating a higher possibility of hallucination. While these approaches have been effective in detecting hallucinated content across various domains, their utilization of multiple forward passes substantially increases computational expense, making them infeasible to apply in real-time or large-scale scenarios. Single-pass methods, such as linear probes on internal states \cite{kossen2024semanticentropyprobesrobust, hewitt-manning-2019-structural}, offer a less computationally intensive solution by making predictions directly from the extracted hidden states of LLMs. These methods avoid the overhead of multiple model generations by leveraging learned representations of uncertainty in a single pass. While state-of-the-art within the context of single-pass methods, linear probes on hidden states are often less effective than multi-pass approaches for hallucination detection. We hypothesize that this performance gap occurs because these probes typically operate on model states at specific points in an LLM's output generation rather than capturing holistic, sequence-level uncertainty patterns and temporal relationships.

To bridge these gaps, we introduce ShED-HD, a novel Shannon-entropy-based hallucination detection framework that maintains high accuracy and generalizability while retaining the efficiency of single-pass approaches. Our approach analyzes token-level entropy distributions using a BiLSTM architecture with single-headed attention. By identifying characteristic uncertainty patterns in output sequences, our model supports robust hallucination detection without the need for multiple model generations. Its lightweight nature also ensures feasibility of real-time deployment on edge-computing devices, where multi-pass methods would be computationally impractical. We conduct extensive experiments on three question-answer datasets: BioASQ \cite{Tsatsaronis2015}, TriviaQA \cite{joshi-etal-2017-triviaqa}, and Jeopardy Questions \cite{kaggle200000Jeopardy} to evaluate ShED-HD's performance, allowing us to evaluate the generalizability of our approach. Our results demonstrate that ShED-HD performance largely outperforms other computationally efficient hallucination detection methods on out-of-distribution data with similar performance on in-distribution data.

Finally, ShED-HD sets the stage for safer and more trustworthy AI-power edge-deployed applications that mitigate the dangers posed by LLM hallucinations and reinforce confidence in their use across a wide range of domains.

\subsection{Key Contributions}
\begin{itemize}
\item We introduce ShED-HD, a novel Shannon-entropy-based hallucination detection framework that captures sequence-level uncertainty patterns while maintaining computational efficiency.
\item We propose a BiLSTM-based architecture with single-headed attention to analyze token-level entropy distributions and detect model hallucination.
\item We demonstrate that ShED-HD outperforms existing computationally efficient hallucination detection methods, particularly in out-of-distribution scenarios.
\item We show that ShED-HD's lightweight design enables real-time inference on edge-computing platforms, making it viable for deployment in resource-constrained environments.
\end{itemize}

\section{Related Work}
\label{sec:related_work}
\subsection{LLM Hallucination}
The challenges of LLM hallucination arise from the fact that responses represent learned probabilities of word occurrences in context rather than guaranteed factual grounding. Model performance correlates with data availability and struggles with obscure or controversial topics where data is limited or highly variable\cite{waldo2024hallucinate}. To predict model hallucinations, researchers have tested a variety of methods outlined below.
\subsection{Retrieval Based Methods}
Min et al. (2023) and Huang and Chen (2024) both introduced retrieval-augmented frameworks for factuality assessment in LLM outputs \cite{min2023factscore} \cite{huang-chen-2024-factalign}. However, one major issue with retrieval-based methods is highlighted in Lee et al. (2024): Despite claims of effective knowledge integration, many models struggle with consistent information retrieval in context-heavy scenarios and often fail to accurately ground their outputs when faced with complex or ambiguous queries \cite{lee-etal-2024-well}. Retrieval-augmented generation methods also introduce substantial computational overhead and may require access to external databases or the internet, reducing practicality in resource-constrained environments \cite{lewis2020retrieval}.
\subsection{Uncertainty-Based Methods}
Uncertainty-based methods have emerged as an efficient approach for hallucination detection in large language models, operating without external knowledge sources or expensive sampling procedures. Early work by Manakul et al. (2023) \cite{manakul2023selfcheckgpt} established that hallucination likelihood correlates with token probability, proposing SelfCheckGPT to leverage this relationship. Huang et al. (2023) \cite{huang2023look} explored various uncertainty metrics for LLMs, demonstrating their effectiveness in identifying potentially hallucinated content. However, these approaches primarily analyze tokens independently, neglecting contextual patterns. To combat this, Zhang et al. (2023) ~\cite{zhang2023enhancing} introduced FOCUS, which addresses co-occurrence bias by propagating uncertainty between tokens. Similarly, Sriramanan et al. (2024) investigate maximal token-level entropy metrics across windows of a model's output sequence as a means for hallucination detection \cite{sriramanan2024llmcheck}. These approaches effectively incorporate an additional degree of context between tokens, but still do not capture temporal context and relations across the entire output sequence.
\subsection{Sampling-Based Methods}
Following this, studies have explored semantic entropy as a means to detect hallucinations by sampling multiple responses from the model and then clustering them algorithmically based on bidirectional entailment—whether one response implies the other and vice versa \cite{farquharSemanticEntropy, kuhn2023semanticuncertaintylinguisticinvariances}. The discrete variant of semantic entropy used by Kossen et al. (2024) \cite{kossen2024semanticentropyprobesrobust}, which is compatible with black-box models, is defined as:
\begin{equation}
H_{\text{SE}}(x) := -\sum_{k=1}^{K} p(C_k|x) \log p(C_k|x).
\end{equation}
where:
\begin{equation}
p(C_k|x) = \sum_{j=1}^{N} \mathbbm{1}[s_j \in C_k]/K
\end{equation}
$(C_1,...,C_K)$ are semantic clusters, and $x$ is the input context. The semantic entropy $H_{\text{SE}}$ is binarized via a threshold to generate hallucination labels. We include semantic entropy as a benchmark.
\subsection{Probe-Based Methods}
Accuracy probes, which predict hallucination labels directly based on the hidden states of a single model generation, have been explored as an efficient means for hallucination detection \cite{alnuhait2024factcheckmatepreemptivelydetectingmitigating, ji2024llminternalstatesreveal}.
Kossen et al. (2024) extend this concept with Semantic Entropy Probes (SEPs), which operate on the hidden states of a single model generation to predict binarized discrete semantic entropy, avoiding the computational overhead of sampling-based semantic uncertainty estimation. \cite{kossen2024semanticentropyprobesrobust}. 

\section{Methods}
\subsection{Data Generation}
We use three distinct datasets to train our pipeline: BioASQ \cite{Tsatsaronis2015}, TriviaQA \cite{joshi-etal-2017-triviaqa}, and Jeopardy Questions \cite{kaggle200000Jeopardy}. We randomly sampled independent train and test splits from each of these datasets of the following sizes: \\

 \begin{table}[h]
    \centering
    \caption{Number of questions in each dataset for train and test.}
    \begin{tabular}{lcc}
        \hline
        \textbf{Dataset} & \textbf{Train Questions} & \textbf{Test Questions} \\
        \hline
        BioASQ & 2,297 & 575 \\
        TriviaQA & 2,500 & 600 \\
        Jeopardy Questions & 2,575 & 600 \\
        \hline
    \end{tabular}
    
    \label{tab:dataset_sizes}
\end{table}

For each question in the train datasets, we generate 3 sample responses. For each question in the test datasets, we generate 10 sample responses to accurately benchmark against multi-pass techniques such as semantic entropy. All response generation is executed by Llama-3.2-1B-Instruct with temperature 0.7 \cite{grattafiori2024llama3herdmodels}. Using Llama-3.2-1B-Instruct is practical due to its small architecture and optimization for CPU inference on edge devices, making it particularly suitable for our target deployment environment. The architectural efficiency of this model allows for reasonable inference times even on hardware with limited computational capabilities, an important consideration for edge computing scenarios. Generating multiple responses per prompt allows us to examine the variability in the model's outputs, providing valuable insights into both response diversity and consistency across different question types and knowledge domains.

To quantify the model's confidence during text generation, we compute a Shannon entropy distribution for each of the model responses. The process works as follows: for each output token—prior to its generation—we calculate the Shannon entropy of the probability distribution over the top 100 most likely tokens. This approach provides a fine-grained measure of uncertainty at each step of the generation process, capturing subtle variations in model confidence that may indicate potential hallucination risk.

Let \( p(i, t) \) represent the probability of the \( i \)-th most probable token at position \( t \), where \( i \in \{1, \dots, 100\} \). The Shannon entropy \( H(t) \) at each token position \( t \) is defined as:
\begin{equation}
H(t) = - \sum_{i=1}^{100} p(i, t) \log p(i, t)
\end{equation}
where:
\begin{itemize}
    \item \( t \) is the current token position.
    \item \( p(i, t) \) is the probability of the \( i \)-th most probable token at position \( t \), computed from the model's logit probabilities
\end{itemize}

As each token is generated, its entropy value is added to an ordered list. Once the entire output sequence is generated, this list represents the complete Shannon entropy distribution, with its length matching the number of tokens in the model's generated response. This sequence of entropy values captures the temporal uncertainity patterns throughout the generation process, providing a strong signal for patterns associated with hallucinations. We then truncate this list to the first 64 entropy values to standardize our representation. We find this does not lead to a significant performance dip, as the most informative entropy patterns typically manifest within the initial portion of the generated sequence.

To assess whether the model's responses contain hallucinated content, we use GPT-4o to compare each generated output against the reference answers provided in the datasets \cite{openai2024gpt4technicalreport}. The comparison focuses on semantic similarity rather than exact matches, allowing for variation in phrasing while maintaining factual accuracy. This approach acknowledges the diversity in natural language expression while still enforcing factual consistency with the reference answers. If a generated response is found to be semantically similar to the reference answer, it is labeled as not a hallucination (i.e., correct). Conversely, if the generated response diverges semantically from the reference, it is labeled as a hallucination. We focus on finding any instance of the reference answer in the response, allowing for extra information beyond the ground truth in the zero-labeled entries. This automated evaluation allows us to efficiently annotate all generated responses with hallucination labels, forming a crucial part of our downstream analysis.

\begin{figure*}[ht]
    \centering
    \includegraphics[width=0.9\textwidth]{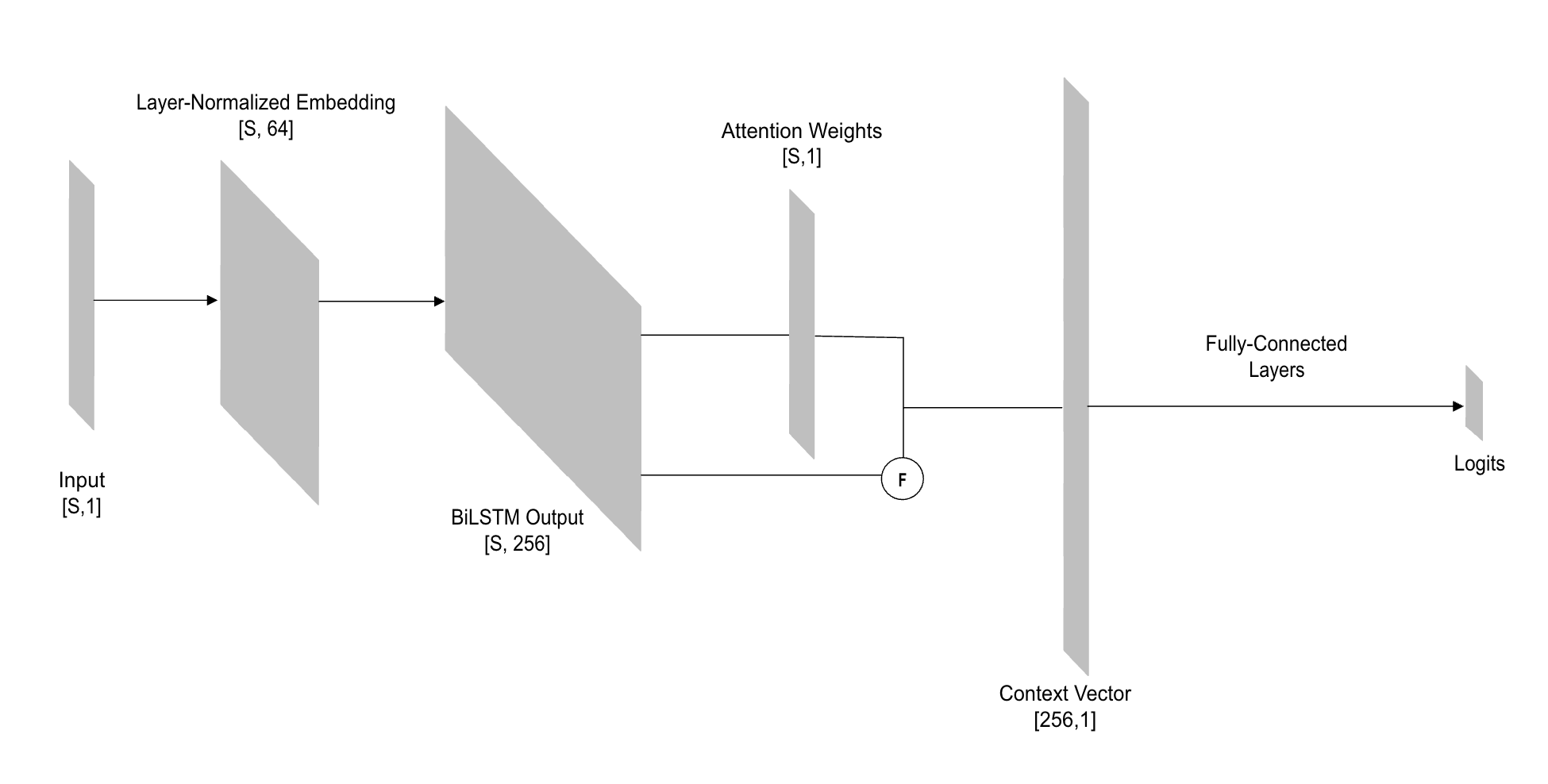}
    \caption{ShED-HD Model Architecture Diagram. $S$ represents the standardized sequence length of a batch, and $F={B^T}A$, where $A$ represents the attention weights and $B$ represents the BiLSTM output. Essentially, $F$ uses the attention weights to compute a weighted row sum on the BiLSTM output.}
    \label{fig:wide}
\end{figure*}

\subsection{Classifier Architecture}
Our classifier is based on a BiLSTM-Attention architecture optimized for sequence-based entropy pattern analysis:
\begin{itemize}
\item \textbf{Input Embedding Layer}: A linear transformation followed by LayerNorm and GELU activation converts each scalar entropy value into a 64-dimensional representation, capturing the features of individual uncertainty measurements. This transformation creates a representation space in which patterns in entropy values can be more effectively processed by subsequent layers.
\item \textbf{BiLSTM}: A 2-layer BiLSTM with 128 hidden units per direction processes the embedded sequence in both forward and backward directions, enabling the model to capture contextual dependencies and temporal patterns in entropy. The bidirectional approach ensures that both preceding and subsequent entropy values influence the representation of each position, providing greater context for uncertainty analysis. The stacked nature of the BiLSTM allows for hierarchical feature extraction, with the second layer building upon the representations learned by the first.
\item \textbf{Attention Mechanism}: An attention module learns importance weights for each position in the sequence, allowing the model to focus on the most informative entropy patterns. This mechanism transforms the variable-length BiLSTM outputs into a fixed-size context vector by computing a weighted row sum. The attention weights themselves provide insights into which parts of the entropy sequence are most important signals for hallucination detection, enhancing both performance and explainability.
\item \textbf{Fully-Connected Layers}: A series of fully-connected layers with BatchNorm, ReLU activations, and dropout refine the context vector for final binary classification.  We use 2 fully-connected layers with hidden dimensions 128 and 64.
\item \textbf{Output Layer}: A linear transformation maps the final hidden representation to a 2-dimensional logit vector, representing the model's confidence scores for the "truth" and "hallucination" classes. These logits are passed through a softmax function during inference to obtain probability estimates, with the higher probability class determining the final prediction.
\end{itemize}
This architecture effectively handles variable-length entropy sequences while focusing on intricate patterns for hallucination detection. The combination of BiLSTM for sequence modeling and attention for selective information aggregation enables the model to identify subtle entropy signatures that distinguish hallucinated from factual content.

\subsection{Training Procedure}
We use a stratified data split (80\% train, 20\% validation) to maintain class distribution across partitions while preventing information leakage between related questions. This ensures that the model is exposed to a representative distribution of hallucinated and non-hallucinated examples during both training and validation.

Training utilizes:
\begin{itemize}
\item \textbf{Optimizer}: AdamW with learning rate $2\times10^{-4}$ and weight decay $2\times10^{-4}$ is used for effective regularization. The adaptive learning rate combined with weight decay helps prevent overfitting while maintaining efficient convergence, particularly important for our relatively small yet complex model architecture.
\item \textbf{Loss Function}: We use a class-weighted cross-entropy loss with weights 1.3 and 0.7, for non-hallucinated and hallucinated classes respectively, to address class imbalance while maintaining sensitivity to hallucinations. These weights were determined through a hyperparameter search to optimize the balance between precision and recall.
\item \textbf{Regularization}: A dropout strategy (0.4) is applied at multiple levels: after embedding, within LSTM layers, in the attention mechanism, and between classifier layers. This further prevents overfitting and improves generalization. The relatively high dropout rate reflects our emphasis on generalization capability, particularly important for cross-domain applications.
\item \textbf{Learning Rate Schedule}: A warm-up strategy is utilized for the first 5 epochs followed by cosine decay, balancing initial training stability with convergence speed. This schedule helps overcome the challenges of training RNN-based models by gradually increasing the learning rate during warm-up and then reducing it to find optimal parameter values.
\item \textbf{Early Stopping}: A patience of 10 epochs is incorporated, monitoring validation macro-F1 score to prevent overfitting while also ensuring model convergence. This balances the trade-off between training time and model performance, allowing sufficient opportunity for improvement while avoiding excessive training.
\item \textbf{Gradient Clipping}: This is applied to prevent exploding gradients, particularly important for LSTM-based architectures.
\end{itemize}

This training procedure is designed to handle imbalanced data while optimizing for precision and recall across both hallucinated and non-hallucinated classes, as well as generalizability.

\begin{table}[h]
    \centering
    \caption{Breakdown of Model Parameters by Component}
    \begin{tabular}{lc}
        \toprule
        \textbf{Model Component} & \textbf{Number of Parameters} \\
        \midrule
        Input Embedding Layer & 256 \\
        BiLSTM  & 593,920 \\
        Attention Mechanism  & 16,513 \\
        Fully-Connected Layers  & 41,536 \\
        Output Layer  & 130 \\
        \midrule
        \textbf{Total Parameters} & \textbf{652,355} \\
        \bottomrule
    \end{tabular}
    \label{tab:model_params}
\end{table}

\section{Evaluation and Results}
\subsection{Performance Metrics}
We use Macro-F1 score as a metric to evaluate classification performance. Macro-F1 is the arithmetic mean of positive and negative class F1 scores. Explicitly, this can be defined as follows: \\
\begin{equation}
\text{Macro-}F_1 = \frac{1}{|T|} \sum_{t \in T} \frac{2 P_t R_t}{P_t + R_t}
\end{equation} \\

where $T$ is the set of all classes, and $P_t$ and $R_t$ and the precision and recall for each class respectively. This balances precision and recall for both the hallucination and non-hallucination cases, giving unbiased insight into the model's performance regardless of class imbalance in the test data. This is especially crucial in applications where both missing hallucinations and incorrectly flagging valid content could have negative consequences.

\begin{figure*}[h] 
    \centering
    \begin{minipage}{0.49\textwidth}
        \centering
        \includegraphics[width=\textwidth]{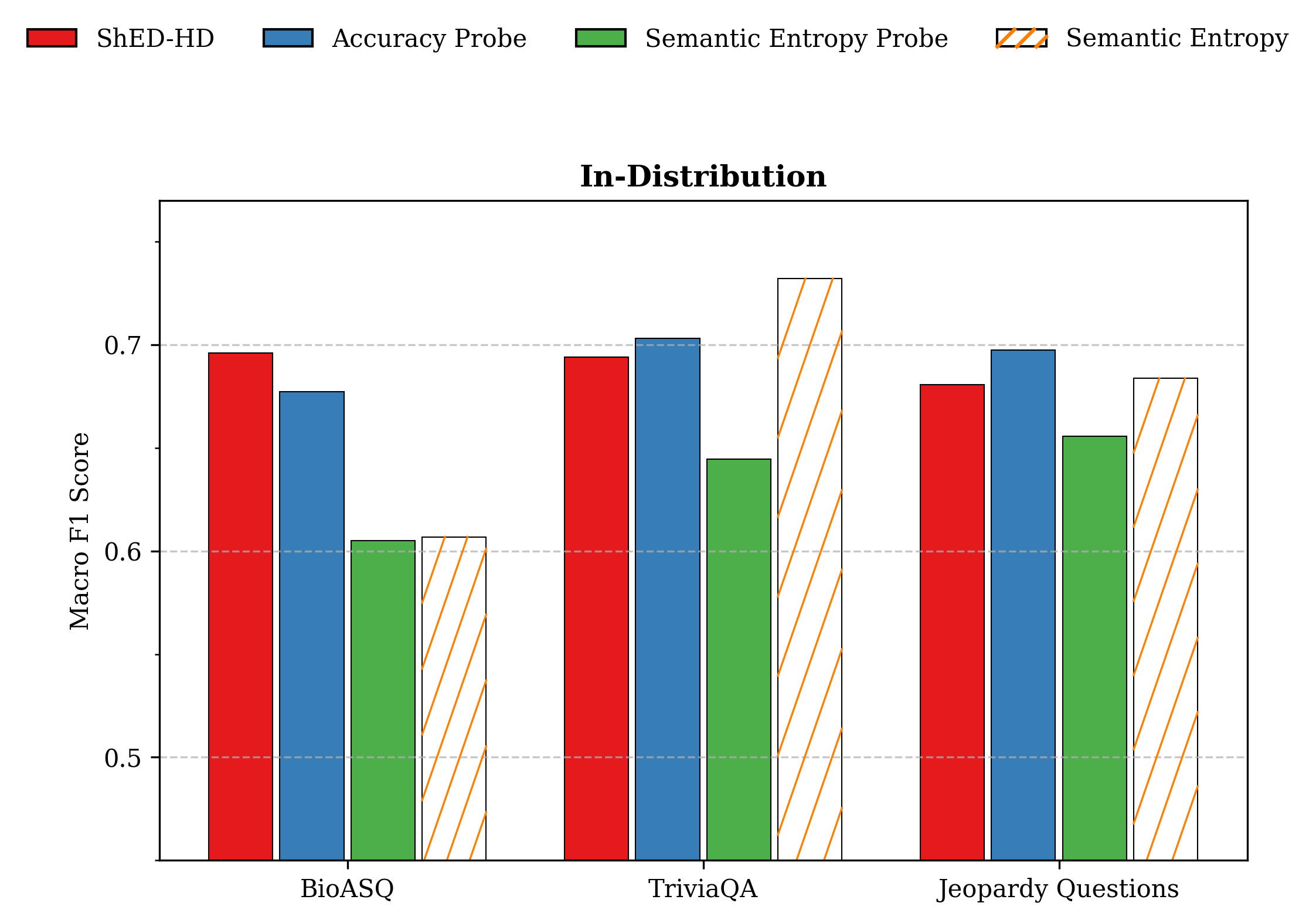}
        \caption{In-Distribution Macro-F1 Results.}
        \label{fig:in_dist}
    \end{minipage}
    \hfill
    \begin{minipage}{0.49\textwidth}
        \centering
        \includegraphics[width=\textwidth]{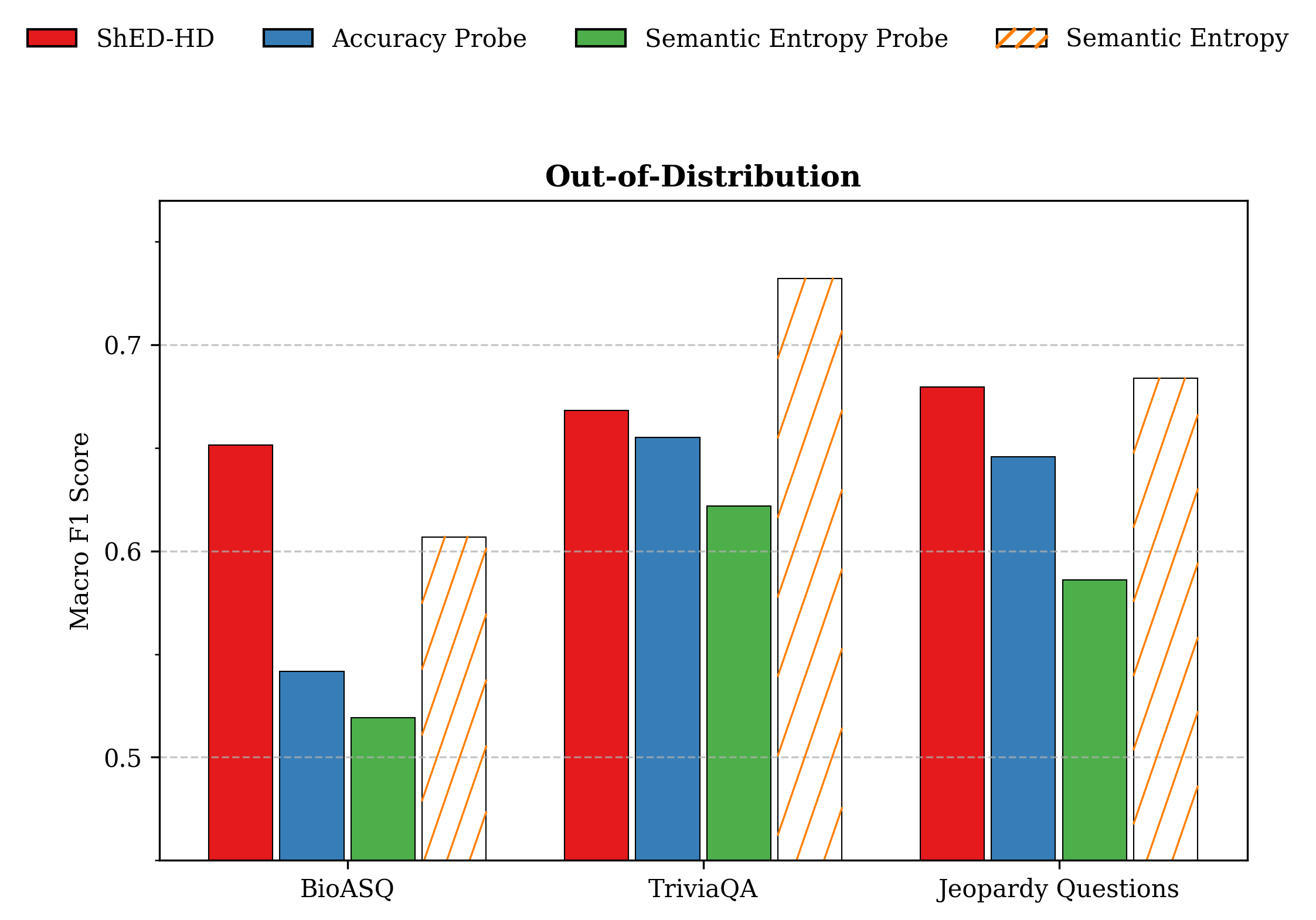}
        \caption{Out-of-Distribution Macro-F1 Results.}
        \label{fig:out_dist}
    \end{minipage}
\end{figure*}

\subsection{Benchmarking}
We benchmark the discrete semantic entropy calculation following Farquhar et al. (2024) \cite{farquharSemanticEntropy}. This approach measures uncertainty at the semantic level rather than the token level, providing a nuanced perspective on model confidence. For each question in our test sets, we sample N=10 responses from Llama-3.2-1B-Instruct. This sampling process allows us to assess the diversity of responses when the model is presented with the same question multiple times, and also allows us to easily calculate discrete semantic entropy for downstream comparison.

Using a DeBERTa-v2-xlarge-mnli natural language inference model, we determine semantic equivalence between text pairs through bi-directional entailment checks \cite{he2021debertadecodingenhancedbertdisentangled}. This approach captures the semantic similarity between responses more effectively than exact matching, allowing us to identify when the model is generating semantically consistent answers versus inconsistent ones. We convert continuous semantic entropy scores into binary labels using an optimal threshold $\gamma$* following Kossen et al. (2024), which results in a binary hallucination label for each response \cite{kossen2024semanticentropyprobesrobust}.

Additionally, we benchmark against accuracy probes and semantic entropy probes, which represent state-of-the-art methods for single-pass hallucination detection methodologies. We train probes on the last hidden state of Llama-3.2-1B-Instruct at the second-to-last token position, meaning the token generated before the stop token, which yields results that strongly trended towards optimal as shown by Kossen et al. (2024) \cite{kossen2024semanticentropyprobesrobust}. This comparison allows us to directly assess the performance improvements of ShED-HD relative to existing methods that operate on single hidden states.

\section{Discussion}
\label{sec:discussion}

\subsection{Comparative Performance Analysis}
Our experimental results confirm the effectiveness of ShED-HD as a generalizable one-pass hallucination detection system compared to state-of-the-art approaches. The performance metrics across the three datasets reflects important trends in the behavior of various techniques for hallucination detection. In out-of-distribution settings, ShED-HD shows a substantial improvement over both accuracy probes and semantic entropy probes. On the BioASQ dataset, for instance, ShED-HD retains macro-F1 of 0.6513, while accuracy probes fall to 0.5416 and semantic entropy probes to 0.5192. This remarkable performance difference underscores ShED-HD's better generalization capabilities, a critical factor in real deployment where domain shifts are unavoidable and models must maintain reliability regardless of changing content type.

Upon comparing in-distribution performance, we observe that ShED-HD competes with or outperforms state-of-the-art techniques like accuracy probes, particularly in the case of the BioASQ data where ShED-HD produces a macro-F1 score of 0.6960 whereas that of the accuracy probe is 0.6772. Competitive in-distribution performance accompanied by significantly improved out-of-distribution performance further emphasizes the robustness of our technique. The ability to maintain consistent performance across domains is especially valuable in practical applications where the deployment environment may differ significantly from the training distribution.

In both the out-of-distribution and in-distribution cases, the comparison with semantic entropy is particularly instructive. While semantic entropy performs well across datasets, it requires multiple forward passes, and this introduces immense computational overhead that may be too large in resource-constrained environments. Each additional inference pass adds computational overhead and latency, rendering this strategy infeasible for real-time use on the edge. ShED-HD's competitive performance with orders of magnitude reduction in computational demand is a major step towards the possibility of practical deployment, with the ability to perform hallucination detection efficiently without sacrificing responsiveness.

The performance gap between ShED-HD and semantic entropy is quite small on datasets like Jeopardy, indicating that our approach picks up many of the same uncertainty signals generated by multi-pass approaches. This insight indicates that temporal trends in the entropy distributions are rich with information that can be effectively leveraged without resorting to computationally expensive sampling procedures.

\subsection{Entropy Pattern Analysis}
The effectiveness of ShED-HD can be attributed to its ability to identify characteristic patterns in token-level entropy distributions that distinguish hallucinated from factual content. Unlike previous approaches that focus on isolated token states, our BiLSTM architecture with single-headed attention processes the entire sequence of entropy values, capturing the most important temporal dependencies and uncertainty fluctuations throughout the generation process. This holistic view allows the model to detect subtle patterns that may be missed by methods that examine only individual tokens, aggregate statistics, or model states at specific points during output generation.

Analysis of the attention weights reveals that ShED-HD focuses on specific regions within the entropy sequences that are particularly indicative of hallucination. The attention mechanism appears to assign greater importance to entropy values from early portions of generated responses as demonstrated in Figures 1 and 2, suggesting that key markers of hallucination emerge during relatively earlier phases of text generation. This finding aligns with the intuition that uncertainty patterns established early in the generation process set the trajectory for the entire response.

\begin{figure*}[h] 
    \centering
    \begin{minipage}{0.49\textwidth}
        \centering
        \includegraphics[width=\textwidth]{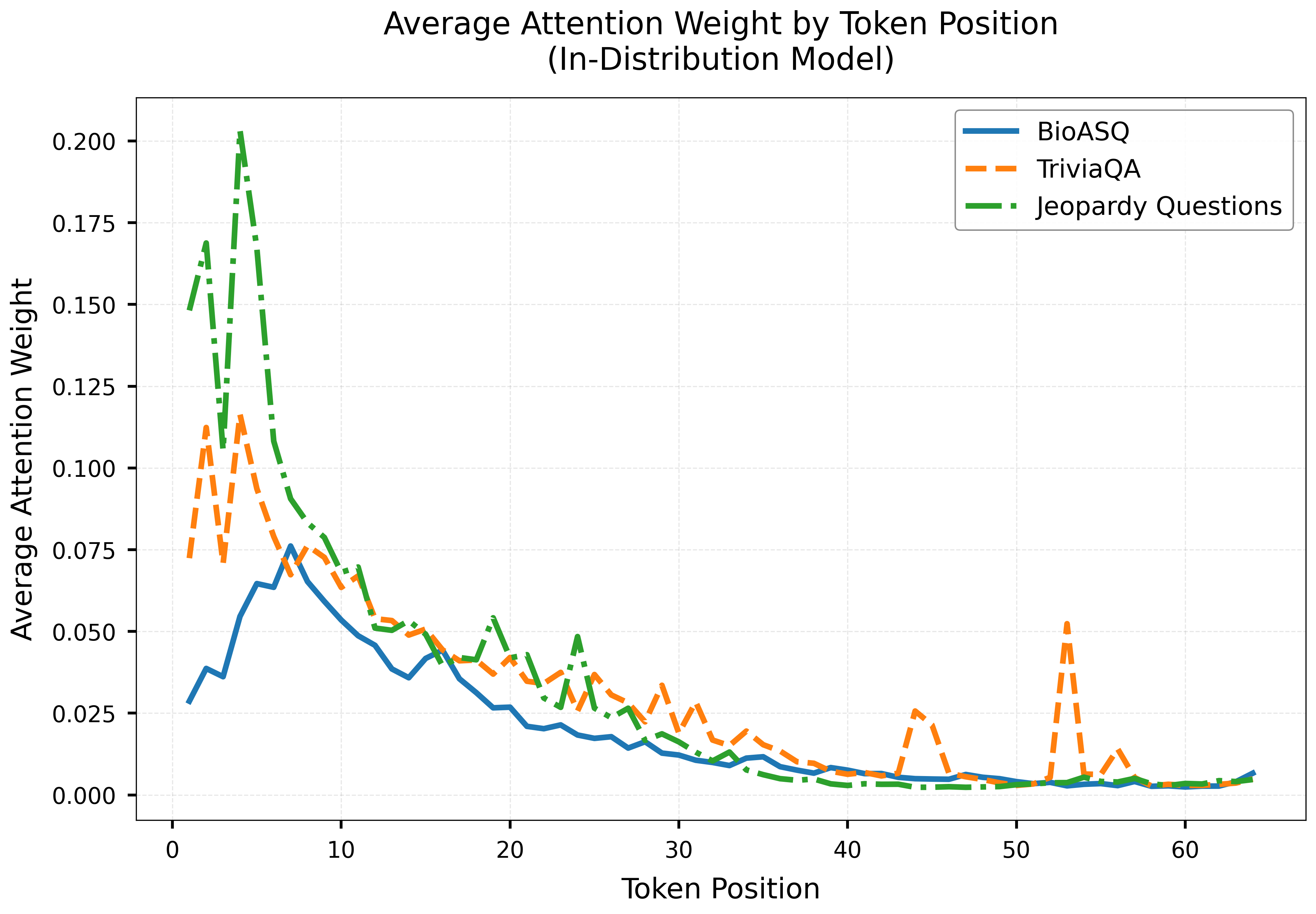}
        \caption{In-Distribution Average Attention Weight vs. Output Token Position.}
        \label{fig:in_dist}
    \end{minipage}
    \hfill
    \begin{minipage}{0.49\textwidth}
        \centering
        \includegraphics[width=\textwidth]{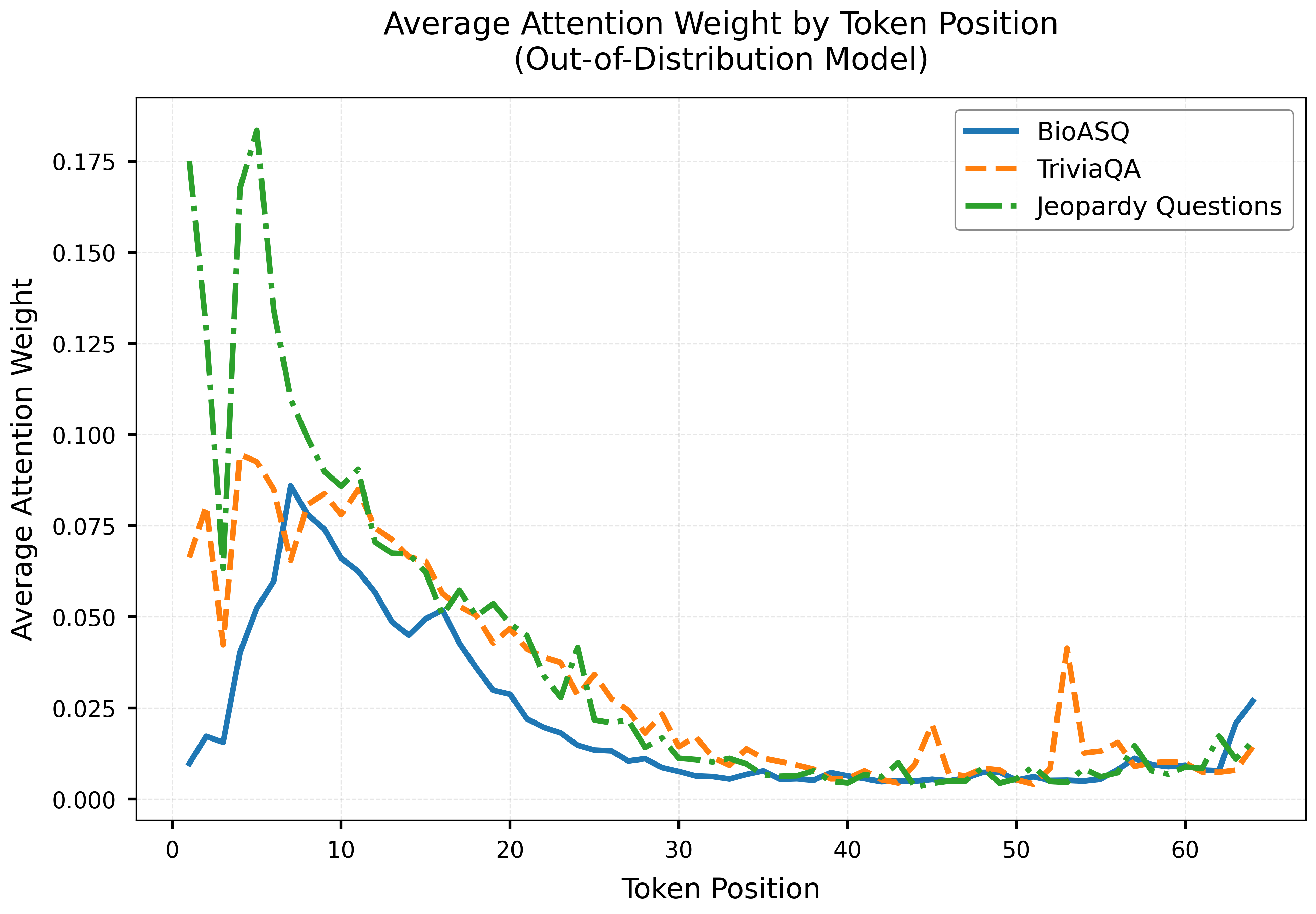}
        \caption{Out-of-Distribution Average Attention Weight vs. Output Token Position.}
        \label{fig:out_dist}
    \end{minipage}
\end{figure*}

\begin{table*}[h]
    \centering
    \begin{minipage}{0.49\textwidth} 
        \centering
        \caption{In-Distribution Macro-F1 results.}
        \small 
        \resizebox{\linewidth}{!}{ 
        \begin{tabular}{lccc}
        \toprule
        \textbf{Method} & \textbf{BioASQ} & \textbf{TriviaQA} & \textbf{Jeopardy Questions} \\
        \midrule
        \textbf{ShED-HD} & 0.6960 & 0.6941 & 0.6806 \\
        \textbf{Accuracy Probe} & 0.6772 & 0.7031 & 0.6976 \\
        \textbf{Semantic Entropy Probe} & 0.6052 & 0.6445 & 0.6556 \\
        \textbf{Semantic Entropy} & 0.6069 & 0.7322 & 0.6838 \\
        \bottomrule
        \end{tabular}
        }
    \end{minipage}
    \hfill
    \begin{minipage}{0.49\textwidth} 
        \centering
        \caption{Out-of-distribution Macro-F1 results.}
        \small 
        \resizebox{\linewidth}{!}{ 
        \begin{tabular}{lccc}
        \toprule
        \textbf{Method} & \textbf{BioASQ} & \textbf{TriviaQA} & \textbf{Jeopardy Questions} \\
        \midrule
        \textbf{ShED-HD} & 0.6513 & 0.6683  & 0.6794 \\
        \textbf{Accuracy Probe} & 0.5416 & 0.6553 & 0.6456 \\
        \textbf{Semantic Entropy Probe} & 0.5192 & 0.6219 & 0.5860 \\
        \textbf{Semantic Entropy} & 0.6069 & 0.7322  & 0.6838\\
        \bottomrule
        \end{tabular}
        }
    \end{minipage}
\end{table*}

The BiLSTM architecture enables the model to situate each entropy value within the context of both future and past uncertainty patterns. It is this bidirectional context that enables the model to distinguish between localized bursts of uncertainty (which may naturally appear in factual content when covering complex concepts) and the persistent, structured uncertainty patterns characteristic of hallucinated content.

\subsection{Cross-Domain Generalization}
One of the most significant findings of our work is ShED-HD's improved generalization capabilities compared to traditional probing approaches. The performance gap between in-distribution and out-of-distribution settings is substantially smaller for ShED-HD than for accuracy probes and semantic entropy probes, particularly on the specialized BioASQ dataset.

This enhanced cross-domain generalization suggests that the sequence-level entropy patterns identified by ShED-HD represent fundamental indicators of hallucination that go past domain-specific features. While traditional probes may learn domain-specific uncertainty signals that fail to generalize, ShED-HD's focus on sequential patterns appears to capture more universal hallucination signatures.  This property is very valuable in real-world applications, where the environment will be different from the training distribution.

The performance on BioASQ—a dataset with domain-specific medical and biological terminology—is especially noteworthy. Technical content in a particular domain typically poses challenges for hallucination detection since it includes intricate and foreign vocabulary. ShED-HD's robustness in this domain shows that patterns of entropy remain good indicators of hallucination even when content is significantly different.

\subsection{Computational Efficiency Considerations}
A core motivation for developing ShED-HD was to address the computational inefficiency of multi-pass approaches like semantic entropy. Our results demonstrate that this goal has been largely achieved, with ShED-HD providing competitive performance while working within the constraints of a single forward pass. This efficiency makes ShED-HD viable for real-time applications and deployment on edge devices where computational resources are limited.

The lightweight architecture of ShED-HD contributes significantly to its efficiency. With a total of 652,355 parameters, the model remains small while retaining sufficient capacity to capture complex entropy patterns. This architectural efficiency translates directly to practical benefits in terms of inference speed and memory requirements, enabling hallucination detection in scenarios where traditional multi-pass approaches would be prohibitively expensive.

Moreover, ShED-HD's efficiency offers the possibility of incorporation into the generation pipeline itself, potentially enabling real-time hallucination detection and prevention. Detecting likely hallucinations during generation could enable systems to impose interventions such as regeneration, factual grounding, or explicit uncertainty recognition to increase output trustworthiness at no considerable latency increase.

\subsection{Implications for LLM Uncertainty Quantification}
Beyond its application in direct hallucination detection, ShED-HD also contributes to the broader field of uncertainty quantification for language models.The observed patterns in entropy distributions provide insights into how uncertainty is represented during text generation, which can be used to guide future work on model calibration and selective prediction.

The success of sequence-level entropy analysis demonstrates that uncertainty in language models has temporal structure that can be exploited for a variety of uncertainty-aware applications. Such structures are capable of reflecting the model's internal estimation of confidence at generation, providing an insight into the model's "epistemic state" without requiring explicit confidence scores or calibration.

Furthermore, cross-domain consistency of such entropy patterns means that some measures of uncertainty are intrinsic to the generation process and not specific to a particular domain. Such an implication bears significance for generic uncertainty quantification schemes attempting to provide sound estimates of confidence for a broad array of application domains.
\subsection{Limitations and Future Work}

Despite ShED-HD's promising results, several limitations warrant further investigation. First, while our approach effectively captures sequence-level uncertainty patterns, it does not incorporate semantic or contextual information from the generated text itself. Integrating lightweight semantic features alongside entropy distributions could potentially enable more fine-grained hallucination detection without significantly compromising computational efficiency.

Second, the binary classification approach adopted in this study may oversimplify the nuanced nature of hallucinations in practical use. Future work may explore finer-grained classification schemes, distinguishing between different types and severities of hallucinations to provide more informative feedback.

Third, while we demonstrate effectiveness on three diverse datasets, all experiments were conducted using the Llama-3.2-1B-Instruct model. Further research is needed to determine whether the entropy patterns learned by ShED-HD generalize across different model architectures and scales. Since the sequential patterns indicative of hallucination are not directly dependent on the model architecture, they may remain consistent, but comprehensive cross-model validation is necessary.

Finally, our current implementation is directed only towards post-generation hallucination detection. One promising direction for future research is integrating ShED-HD into the generation process itself, which may enable real-time intervention when hallucination patterns are detected. Such integration can lead to self-correcting generation systems that actively inhibit hallucinations during content generation.

\section{Conclusion}
\label{sec:conclusion}

This paper introduces ShED-HD, a novel Shannon-entropy-based hallucination detection framework that achieves robust performance with computational efficiency, rendering it an effective solution for edge-compute scenarios. Through the analysis of sequence-level patterns in token-level entropy distributions by a specialized BiLSTM model with single-headed attention, our approach offers significant performance improvements over the state-of-the-art computationally efficient methods without invoking the overhead of multi-pass approaches.

Our comprehensive evaluation across three datasets (BioASQ, TriviaQA, and Jeopardy Questions) demonstrates ShED-HD's effectiveness in both in-distribution and out-of-distribution settings, with particularly strong cross-domain generalization capabilities. The computational efficiency of ShED-HD makes it suitable for real-time applications and resource-constrained environments where traditional multi-pass approaches would be impractical.

Our comprehensive experiments on three datasets (BioASQ, TriviaQA, and Jeopardy Questions) confirm the effectiveness of ShED-HD in both in-distribution and out-of-distribution settings, with state-of-the-art cross-domain generalization performance. The computational efficiency of ShED-HD enables its application in real-time and computationally restricted environments where traditional multi-pass approaches would be impractical.

As LLMs are used more in high-stakes domains, reliable hallucination detection becomes increasingly crucial. ShED-HD is a significant step toward ensuring the reliability of LLM outputs and meeting practical deployment constraints. Future work will focus on the addition of lightweight semantic features, the exploration of more fine-grained classification schemes, and investigating if learned entropy patterns are transferable across model architectures.

\section{Additional Information}
We prompt Llama-3.2-1B-Instruct as follows to generate responses to questions, following the long-form generation strategy of Kossen et al. (2024)  \cite{kossen2024semanticentropyprobesrobust}:
\begin{verbatim}
Answer the following question in a
single brief but complete sentence.

Question: {question}

Answer:
\end{verbatim}

We prompt GPT-4o as follows to generate hallucination labels for responses:
\begin{verbatim}
We are assessing the factual accuracy 
of answers to the following question:

Question: "{question}"

The factually correct answers are:
{answers_str}

The proposed answer is: "{model_output}"

Is the proposed answer factually correct 
(matches any of the correct answers in 
meaning), or is it a hallucination? A 
hallucination means the answer contains 
factually incorrect information or 
contradicts the correct answers. Respond 
only with "factual" or "hallucination".

Response:
\end{verbatim}

\begin{table}[h]
    \centering
    \caption{Accuracy of Llama-3.2-1B-Instruct on each dataset.}
    \begin{tabular}{lcc}
        \hline
        \textbf{Dataset} & \textbf{Train Accuracy} & \textbf{Test Accuracy} \\
        \hline
        BioASQ & 40.02\% & 41.69\% \\
        TriviaQA & 33.40\% & 33.05\% \\
        Jeopardy Questions & 34.16\% & 34.38\% \\
        \hline
    \end{tabular}
    
    \label{tab:dataset_sizes}
\end{table}

\bibliography{references}

\end{document}